\documentclass[11pt,letterpaper]{article}
\usepackage{emnlp2017}
\usepackage{times}
\usepackage{latexsym}
\usepackage{CJKutf8}
\usepackage{bbm}
\usepackage{dsfont}
\usepackage{qtree}
\usepackage{multirow}
\usepackage{url}
\usepackage{graphicx}
\usepackage{tikz-qtree}
\usepackage{mathtools}

\DeclarePairedDelimiter\floor{\lfloor}{\rfloor}
\usepackage{amsmath,amssymb,amsthm}

\newcommand{\R}{\ensuremath{\mathbb{R}}}

\newcommand{\ra}{\ensuremath{\rightarrow}}

\newcommand{\paren}[1]{\left(#1\right)}

\newcommand{\abs}[1]{\left|#1\right|}

\newcommand{\by}{\ensuremath{\times}}

\emnlpfinalcopy

\def\emnlppaperid{650}

\title{A Sub-Character Architecture for Korean Language Processing}

\author{
  {\bf Karl Stratos} \\
  Toyota Technological Institute at Chicago \\
  {\tt stratos@ttic.edu}
  }

\date{}

\begin{document}
\maketitle
\begin{CJK}{UTF8}{gbsn}

\begin{abstract}

We introduce a novel sub-character architecture that exploits a unique compositional structure of the Korean language.
Our method decomposes each character into a small set of primitive phonetic units called jamo letters from which character- and word-level representations are induced.
The jamo letters divulge syntactic and semantic information that is difficult to access with conventional character-level units.
They greatly alleviate the data sparsity problem, reducing the observation space to 1.6\% of the original while increasing accuracy in our experiments.
We apply our architecture to dependency parsing and achieve dramatic improvement over strong lexical baselines.

\end{abstract}

\section{Introduction}
\label{sec:intro}

Korean is generally recognized as a language isolate: that is, it has no apparent genealogical relationship with other languages \cite{song2006korean,campbell2007glossary}.
A unique feature of the language is that each character is composed of a small, fixed set of basic phonetic units called \textbf{jamo} letters.
Despite the important role jamo plays in encoding syntactic and semantic information of words,
it has been neglected in existing modern Korean processing algorithms.
In this paper, we bridge this gap by introducing a novel compositional neural architecture that explicitly leverages the sub-character information.

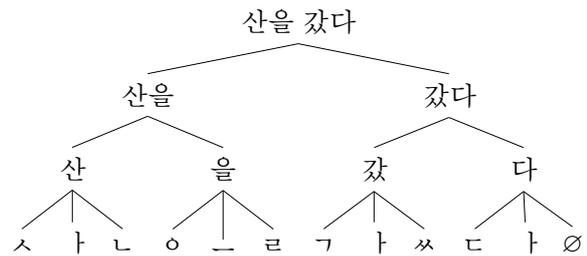
\begin{figure}[t!]
{
\begin{center}
\begin{tikzpicture}[scale=0.93]
  \Tree[.\mbox{\begin{CJK}{UTF8}{mj}산을 갔다\end{CJK}}
    [.\mbox{\begin{CJK}{UTF8}{mj}산을\end{CJK}}
      [.\mbox{\begin{CJK}{UTF8}{mj}산\end{CJK}} \mbox{\begin{CJK}{UTF8}{mj}ㅅ\end{CJK}} \mbox{\begin{CJK}{UTF8}{mj}ㅏ\end{CJK}} \mbox{\begin{CJK}{UTF8}{mj}ㄴ\end{CJK}} ]
      [.\mbox{\begin{CJK}{UTF8}{mj}을\end{CJK}} \mbox{\begin{CJK}{UTF8}{mj}ㅇ\end{CJK}} \mbox{\begin{CJK}{UTF8}{mj}ㅡ\end{CJK}} \mbox{\begin{CJK}{UTF8}{mj}ㄹ\end{CJK}} ]]
    [.\mbox{\begin{CJK}{UTF8}{mj}갔다\end{CJK}}
      [.\mbox{\begin{CJK}{UTF8}{mj}갔\end{CJK}} \mbox{\begin{CJK}{UTF8}{mj}ㄱ\end{CJK}} \mbox{\begin{CJK}{UTF8}{mj}ㅏ\end{CJK}} \mbox{\begin{CJK}{UTF8}{mj}ㅆ\end{CJK}} ]
      [.\mbox{\begin{CJK}{UTF8}{mj}다\end{CJK}} \mbox{\begin{CJK}{UTF8}{mj}ㄷ\end{CJK}} \mbox{\begin{CJK}{UTF8}{mj}ㅏ\end{CJK}} \mbox{$\varnothing$} ]]]
\end{tikzpicture}
\caption{Korean sentence ``\mbox{\begin{CJK}{UTF8}{mj}산을 갔다\end{CJK}}'' (\textit{I went to the mountain}) decomposed to words, characters, and jamos.}
\label{fig:korean}
\end{center}
}
\vspace{-5mm}
\end{figure}

Specifically, we perform Unicode decomposition on each Korean character to recover its underlying jamo letters and construct character- and word-level
representations from these letters.
See Figure~\ref{fig:korean} for an illustration of the decomposition.
The decomposition is \textit{deterministic}; this is a crucial departure from previous work that uses
language-specific sub-character information such as radical (a graphical component of a Chinese character).
The radical structure of a Chinese character does not follow any systematic process,
requiring an incomplete dictionary mapping between characters and radicals to take advantage of this information \citep{sun2014radical,yinmulti}.
In contrast, our Unicode decomposition does not need any supervision and can extract correct jamo letters for all possible Korean characters.

Our jamo architecture is fully general and can be plugged in any Korean processing network.
For a concrete demonstration of its utility, in this work we focus on dependency parsing.
\newcite{mcdonald2013universal} note that ``Korean emerges as a very clear outlier'' in their cross-lingual parsing experiments on the universal treebank,
implying a need to tailor a model for this language isolate.
Because of the compositional morphology, Korean suffers extreme data sparsity at the word level:
2,703 out of 4,698 word types ($> 57\%$) in the held-out portion of our treebank are OOV.
This makes the language challenging for simple lexical parsers even when augmented with a large set of pre-trained word representations.

While such data sparsity can also be alleviated by incorporating more conventional character-level information,
we show that incorporating jamo is an effective and economical new approach to combating the sparsity problem for Korean.
In experiments, we decisively improve the LAS of the lexical BiLSTM parser of \newcite{TACL885} from 82.77 to 91.46
while reducing the size of input space by 98.4\% when we replace words with jamos.
As a point of reference, a strong feature-rich parser using \textit{gold} POS tags obtains 88.61.

To summarize, we make the following contributions.
\begin{itemize}
\item To our knowledge, this is the first work that leverages jamo in end-to-end neural Korean processing.
To this end, we develop a novel sub-character architecture based on deterministic Unicode decomposition.
\item We perform extensive experiments on dependency parsing to verify the utility of the approach.
We show clear performance boost with a drastically smaller set of parameters. Our final model outperforms strong baselines by a large margin.
\item We release an implementation of our jamo architecture which can be plugged in any Korean processing network.\footnote{\url{https://github.com/karlstratos/koreannet}}
\end{itemize}

\section{Related Work}
\label{sec:related-work}

We make a few additional remarks on related work to better situate our work.
Our work follows the successful line of work on incorporating sub-lexical information to neural models.
Various \textit{character}-based architectures have been proposed.
For instance, \newcite{ma-hovy:2016:P16-1} and \newcite{kim2016character} use CNNs over characters whereas
\newcite{lample2016neural} and \newcite{ballesteros:2015emnlp} use bidirectional LSTMs (BiLSTMs).
Both approaches have been shown to be profitable; we employ a BiLSTM-based approach.

Many previous works have also considered \textit{morphemes} to augment lexical models \cite{luong2013better,botha2014compositional,cotterell2016morphological}.
Sub-character models are substantially rarer; an extreme case is considered by
\newcite{gillick2015multilingual} who process text as a sequence of bytes.
We believe that such byte-level models are too general and that there are opportunities to exploit
natural sub-character structure for certain languages such as Korean and Chinese.

There exists a line of work on exploiting graphical components of Chinese characters called radicals \citep{sun2014radical,yinmulti}.
For instance, 足 (\textit{foot}) is the radical of 跑 (\textit{run}).
While related, our work on Korean is distinguished in critical ways and should not be thought of as just an extension to another language.
First, as mentioned earlier, the compositional structure is fundamentally different between Chinese and Korean.
The mapping between radicals and characters in Chinese is nondeterministic and can only be loosely approximated by an incomplete dictionary.
In contrast, the mapping between jamos and Korean characters is deterministic (Section~\ref{subsec:jamo}),
allowing for systematic decomposition of all possible Korean characters.
Second, the previous work on Chinese radicals was concerned with learning word embeddings.
We develop an end-to-end compositional model for a downstream task: parsing.

\section{Method}

\subsection{Jamo Structure of the Korean Language}
\label{subsec:jamo}

Let $\mathcal{W}$ denote the set of word types and $\mathcal{C}$ the set of character types.
In many languages, $c \in \mathcal{C}$ is the most basic unit that is meaningful.
In Korean, each character is further composed of a small fixed set of phonetic units called jamo letters $\mathcal{J}$ where $\abs{\mathcal{J}} = 51$.
The jamo letters are categorized as head consonants $\mathcal{J}_h$, vowels $\mathcal{J}_v$, or tail consonants $\mathcal{J}_t$.
The composition is completely systematic.
Given any character $c \in \mathcal{C}$, there exist $c_h \in \mathcal{J}_h$, $c_v \in \mathcal{J}_v$, and $c_t \in \mathcal{J}_t$ such that their composition yields $c$.
Conversely, any $c_h \in \mathcal{J}_h$, $c_v \in \mathcal{J}_v$, and $c_t \in \mathcal{J}_t$ can be composed to yield a valid character $c \in \mathcal{C}$.

As an example, consider the word \begin{CJK}{UTF8}{mj}갔다\end{CJK} (\texttt{went}).
It is composed of two characters, $\mbox{\begin{CJK}{UTF8}{mj}갔\end{CJK}}, \mbox{\begin{CJK}{UTF8}{mj}다\end{CJK}} \in \mathcal{C}$.
Each character is furthermore composed of three jamo letters as follows:
\begin{itemize}
\item $\mbox{\begin{CJK}{UTF8}{mj}갔\end{CJK}}  \in \mathcal{C}$  is composed of $\mbox{\begin{CJK}{UTF8}{mj}ㄱ\end{CJK}} \in \mathcal{J}_h$, $\mbox{\begin{CJK}{UTF8}{mj}ㅏ\end{CJK}} \in \mathcal{J}_v$, and $\mbox{\begin{CJK}{UTF8}{mj}ㅆ\end{CJK}} \in \mathcal{J}_t$.
\item $\mbox{\begin{CJK}{UTF8}{mj}다\end{CJK}} \in \mathcal{C}$ is composed of $\mbox{\begin{CJK}{UTF8}{mj}ㄷ\end{CJK}}  \in \mathcal{J}_h$, $\mbox{\begin{CJK}{UTF8}{mj}ㅏ\end{CJK}} \in \mathcal{J}_v$, and an empty letter $\mbox{$\varnothing$} \in \mathcal{J}_t$.
\end{itemize}
The tail consonant can be empty; we assume a special symbol $\mbox{$\varnothing$} \in \mathcal{J}_t$ to denote an empty letter.
Figure~\ref{fig:korean} illustrates the decomposition of a Korean sentence down to jamo letters.

Note that the number of possible characters is combinatorial in the number of jamo letters, loosely upper bounded by $51^3 = 132,651$.
This upper bound is loose because certain combinations are invalid.
For instance, $\mbox{\begin{CJK}{UTF8}{mj}ㅁ\end{CJK}} \in \mathcal{J}_h \cap \mathcal{J}_t$
but $\mbox{\begin{CJK}{UTF8}{mj}ㅁ\end{CJK}} \not\in \mathcal{J}_v$
whereas $\mbox{\begin{CJK}{UTF8}{mj}ㅏ\end{CJK}} \in \mathcal{J}_v$ but $\mbox{\begin{CJK}{UTF8}{mj}ㅏ\end{CJK}} \not\in \mathcal{J}_h \cup \mathcal{J}_t$.

The combinatorial nature of Korean characters motivates the compositional architecture below.
For completeness, we describe the entire forward pass of the transition-based BiLSTM parser of \newcite{TACL885} that we use in our experiments.

\subsection{Jamo Architecture}

The parameters associated with the jamo layer are
\begin{itemize}
\item Embedding $e^l \in \R^d$ for each letter $l \in \mathcal{J}$
\item $U^{\mathcal{J}}, V^{\mathcal{J}}, W^{\mathcal{J}} \in \R^{d \by d}$ and $b^{\mathcal{J}} \in \R^d$
\end{itemize}
Given a Korean character $c \in \mathcal{C}$, we perform Unicode decomposition (Section~\ref{subsec:unicode}) to
recover the underlying jamo letters $c_h, c_v, c_t \in \mathcal{J}$.
We compose the letters to induce a representation of $c$ as
\begin{align*}
h^c &= \tanh\paren{U^{\mathcal{J}} e^{c_h} + V^{\mathcal{J}} e^{c_v} + W^{\mathcal{J}} e^{c_t} + b^{\mathcal{J}}}
\end{align*}
This representation is then concatenated with a character-level lookup embedding,
and the result is fed into an LSTM to produce a word representation.
We use an LSTM \cite{hochreiter1997long} simply as a mapping $\phi:\R^{d_1} \times \R^{d_2} \ra \R^{d_2}$
that takes an input vector $x$ and a state vector $h$ to output a new state vector $h' = \phi(x, h)$.
The parameters associated with this layer are
\begin{itemize}
\item Embedding $e^c \in \R^{d'}$ for each $c \in \mathcal{C}$
\item Forward LSTM $\phi^f: \R^{d + d'} \times \R^d \ra \R^d$
\item Backward LSTM $\phi^b: \R^{d + d'}\times \R^d  \ra \R^d$
\item $U^{\mathcal{C}} \in \R^{d \by 2 d}$ and $b^{\mathcal{C}} \in \R^d$
\end{itemize}
Given a word $w \in \mathcal{W}$ and its character sequence $c_1 \ldots c_m \in \mathcal{C}$,
we compute
\begin{align*}
f^c_i &= \phi^f\paren{ \begin{bmatrix}h^{c_i} \\ e^{c_i} \end{bmatrix}, f^c_{i-1} } &&\forall i=1\ldots m \\
b^c_i &= \phi^b\paren{ \begin{bmatrix}h^{c_i}  \\ e^{c_i} \end{bmatrix}, b^c_{i+1} } &&\forall i=m\ldots 1
\end{align*}
and induce a representation of $w$ as
\begin{align*}
h^w &= \tanh{\paren{U^{\mathcal{C}} \begin{bmatrix}f^c_m\\ b^c_1 \end{bmatrix} + b^{\mathcal{C}}}}
\end{align*}
Lastly, this representation is concatenated with a word-level lookup embedding (which can be initialized with pre-trained word embeddings),
and the result is fed into a BiLSTM network.
The parameters associated with this layer are
\begin{itemize}
\item Embedding $e^w \in \R^{d_{\mathcal{W}}}$ for each $w \in \mathcal{W}$
\item Two-layer BiLSTM $\Phi$ that maps $h_1 \ldots h_n \in \R^{d + d_{\mathcal{W}}}$ to $z_1 \ldots z_n \in \R^{d^*}$
\item Feedforward for predicting transitions
\end{itemize}
Given a sentence $w_1 \ldots w_n \in \mathcal{W}$, the final $d^*$-dimensional word representations are given by
\begin{align*}
(z_1 \ldots z_n) &= \Phi\paren{\begin{bmatrix}h^{w_1}\\ e^{w_1} \end{bmatrix} \ldots \begin{bmatrix}h^{w_n}\\ e^{w_n} \end{bmatrix}}
\end{align*}
The parser then uses the feedforward network to greedily predict transitions based on words that are active in the system.
The model is trained end-to-end by optimizing a max-margin objective.
Since this part is not a contribution of this paper, we refer to \newcite{TACL885} for details.

By setting the embedding dimension of jamos $d$, characters $d'$, or words $d_{\mathcal{W}}$ to zero,
we can configure the network to use any combination of these units.
We report these experiments in Section~\ref{sec:experiments}.

\subsection{Unicode Decomposition}
\label{subsec:unicode}

Our architecture requires dynamically extracting jamo letters given any Korean character.
This is achieved by simple Unicode manipulation.
For any Korean character $c \in C$ with Unicode value $U(c)$,
let $\overline{U}(c) = U(c) - 44032$ and $T(c) = \overline{U}(c) \bmod 28$.
Then the Unicode values $U(c_h)$, $U(c_v)$, and $U(c_t)$ corresponding to the head consonant, vowel, and tail consonant are obtained by

\vspace{-3mm}
{\small
\begin{align*}
U(c_h) &= 1 + \floor*{\frac{\overline{U}(c)}{588}} + \texttt{0x10ff} \\
U(c_v) &= 1 + \floor*{\frac{(\overline{U}(c) - T(c)) \bmod 588}{28}} + \texttt{0x1160}\\
U(c_t) &= 1 + T(c) + \texttt{0x11a7}
\end{align*}
}
\vspace{-3mm}

\noindent
where $c_t$ is set to $\mbox{$\varnothing$}$ if $T(c_t) = 0$.

\subsection{Why Use Jamo Letters?}

The most obvious benefit of using jamo letters is alleviating data sparsity by flattening the combinatorial space of Korean characters.
We discuss some additional explicit benefits.
First, jamo letters often indicate syntactic properties of words.
For example, a tail consonant \begin{CJK}{UTF8}{mj}ㅆ\end{CJK} strongly implies that the word is a past tense verb
as in \begin{CJK}{UTF8}{mj}갔다\end{CJK} (\texttt{went}), \begin{CJK}{UTF8}{mj}왔다\end{CJK} (\texttt{came}), and \begin{CJK}{UTF8}{mj}했다\end{CJK} (\texttt{did}).
      Thus a jamo-level model can identify unseen verbs more effectively than word- or character-level models.
Second, jamo letters dictate the sound of a character.
For example, \begin{CJK}{UTF8}{mj}갔\end{CJK} is pronounced as \texttt{got} because the head consonant
\begin{CJK}{UTF8}{mj}ㄱ\end{CJK} is associated with the sound \texttt{g}, the vowel
\begin{CJK}{UTF8}{mj}ㅏ\end{CJK} with \texttt{o}, and the tail consonant
\begin{CJK}{UTF8}{mj}ㅆ\end{CJK} with \texttt{t}.
This is clearly critical for speech recognition/synthesis and indeed has been investigated in the speech community \cite{lee1994phonemie,sakti2010korean}.
While speech processing is not our focus, the phonetic signals can capture useful lexical correlation (e.g., for onomatopoeic words).

\section{Experiments}
\label{sec:experiments}

\paragraph{Data}
We use the publicly available Korean treebank in the universal treebank version 2.0 \cite{mcdonald2013universal}.\footnote{\url{https://github.com/ryanmcd/uni-dep-tb}}
The dataset comes with a train/development/test split; data statistics are shown in Table~\ref{tab:data}.
Since the test portion is significantly smaller than the dev portion, we report performance on both.

\begin{table}[t!]
\begin{center}
{\footnotesize
\begin{tabular}{|c|c|c|c|}
\hline
                          &      Training    &     Development   &   Test \\
\hline
\# projective trees       &      5,425     &     603   &   299  \\
\# non-projective trees   &        12     &       0   &     0  \\
\hline
\end{tabular}

\vspace{2mm}
\begin{tabular}{|c|c|c|c|}
\hline
              &    \#   & \# Ko  &   Examples \\
\hline
word          &      31,060         &      --            &   \begin{CJK}{UTF8}{mj}프로그램보다\end{CJK} \begin{CJK}{UTF8}{mj}갈비   booz\end{CJK}\\
char          &      1,772          &      1,315         &   \begin{CJK}{UTF8}{mj}최 귤 흠 냥 셧 캐 쪽 @ 正 a \end{CJK}\\
jamo          &      500           &       48            &   \begin{CJK}{UTF8}{mj}ㄱ ㄳ ㄼ ㅏ ㅠ ㅢ @ 正 a \end{CJK}\\
\hline
\end{tabular}
\caption{Treebank statistics. Upper: Number of trees in the split.
Lower: Number of unit types in the training portion.
For simplicity, we include non-Korean symbols (e.g., \texttt{@}, \begin{CJK}{UTF8}{mj}正\end{CJK}, \texttt{a}) as characters/jamos.  }
\label{tab:data}
}
\end{center}
\vspace{-4mm}
\end{table}

\begin{table*}[t!]
\begin{center}
{\small
\begin{tabular}{|c|c|c|c|c|c|c|c|c|}
\hline
System            &      Features     &      Feature Representation   &   Emb     &   POS  & \multicolumn{2}{|c|}{Dev}  & \multicolumn{2}{|c|}{Test} \\
                  &                   &                               &           &        & \multicolumn{1}{|c}{UAS}   & \multicolumn{1}{c|}{LAS} &  \multicolumn{1}{|c}{UAS}   & \multicolumn{1}{c|}{LAS} \\
\hline
McDonald13        &   cross-lingual features    &   large sparse matrix       & -- & PRED & --    & -- & 71.22 & 55.85 \\
Yara (beam 64)    &   features in  Z\&N11       &   large sparse matrix       & -- & PRED & 76.31 & 62.83 & 91.19 & 85.17 \\
                  &                             &                             &    & GOLD & 79.08 & 68.85 & 92.93 & 88.61 \\
K\&G16            &    word                     &   $31060 \by 100$  matrix   & -- & --   & 68.87 & 48.25 & 88.61 & 78.95 \\
                  &                             &   $298115 \by 100$  matrix  & YES&      & 76.30 & 60.88 & 90.00 & 82.77 \\
Dyer15            &      word, transition       &   $31067 \by 100$  matrix   & -- & --   & 69.40 & 48.46 & 88.41 & 78.22 \\
                  &                             &   $298122 \by 100$ matrix   & YES&      & 75.99 & 59.38 & 90.73 & 83.89 \\
Ballesteros15     &      char, transition       &   $1779 \by 100$ matrix     & -- & --   & 84.22 & 76.41 & 91.27 & 86.25 \\
\hline
\hline
KoreanNet     &    char        &      $1772 \by 100$  matrix   & --  &  --    &  84.76      & 76.95        & 94.75   & 90.81  \\
              &                &      $1772 \by 200$  matrix   &     &        &  84.83      & 77.29        & 94.55   & 91.04 \\
              &    jamo        &      $500 \by 100$  matrix    & --  &  --    &  84.27	 & 76.07        & 94.59   & 90.77 \\
              &                &      $500 \by 200$  matrix    &     &        &  84.68      & 77.27        & 94.86   & 91.46 \\
              &    char, jamo  &      $2272 \by 100$  matrix   & --  &  --    &  85.35	 & 78.18        & 94.79   & 91.19 \\
              &                &      $2272 \by 200$  matrix   &     &        &  85.74	 & 78.76        & 94.55   & 91.31 \\
              & word, char, jamo  &      $302339 \by 200$  matrix & YES &  --    & \textbf{86.39}      & \textbf{79.68}        & \textbf{95.17}   & \textbf{92.31} \\
\hline
\end{tabular}
\caption{Main result. Upper: Accuracy with baseline models. Lower: Accuracy with different configurations of our parser network (word-only is identical to K\&G16).}
\label{tab:result}
}
\end{center}
\vspace{-3mm}
\end{table*}

As expected, we observe severe data sparsity with words: 24,814 out of 31,060 elements in the vocabulary appear only \textit{once} in the training data.
On the dev set, about $57\%$ word types and $3\%$ character types are OOV.
Upon Unicode decomposition, we obtain the following 48 jamo types:

\vspace{1mm}
\begin{center}
{
\fbox{\parbox{2.6in}{\begin{CJK}{UTF8}{mj}ㄱ ㄳ ㄲ ㄵ ㄴ ㄷ ㄶ ㄹ ㄸ ㄻ ㄺ ㄼ ㅁ ㅀ ㅃ ㅂ ㅅ ㅄ ㅇ ㅆ ㅉ ㅈ ㅋ ㅊ ㅍ ㅌ ㅏ ㅎ ㅑ ㅐ ㅓ ㅒ ㅕ ㅔ ㅗ ㅖ ㅙ ㅘ ㅛ ㅚ ㅝ ㅜ ㅟ ㅞ ㅡ ㅠ ㅣ ㅢ \end{CJK}}}
}
\end{center}
\vspace{1mm}

\noindent
none of which is OOV in the dev set.

\paragraph{Implementation and baselines}
We implement our jamo architecture using the DyNet library \cite{dynet} and plug it into the BiLSTM parser of \newcite{TACL885}.\footnote{\url{https://github.com/elikip/bist-parser}}
For Korean syllable manipulation, we use the freely available toolkit by Joshua Dong.\footnote{\url{https://github.com/JDongian/python-jamo}}
We train the parser for 30 epochs and use the dev portion for model selection.
We compare our approach to the following baselines:
\begin{itemize}
\item McDonald13: A cross-lingual parser originally reported in \newcite{mcdonald2013universal}.
\item Yara: A beam-search transition-based parser of \newcite{DBLP:journals/corr/RasooliT15} based on the rich non-local features in \newcite{zhang2011transition}.
We use beam width 64. We use 5-fold jackknifing on the training portion to provide POS tag features. We also report on using \textit{gold} POS tags.
\item K\&G16: The basic BiLSTM parser of \newcite{TACL885} without the sub-lexical architecture introduced in this work.
\item Stack LSTM: A greedy transition-based parser based on stack LSTM representations.
  Dyer15 denotes the word-level variant \cite{dyer:2015acl}.
  Ballesteros15 denotes the character-level variant \cite{ballesteros:2015emnlp}.
\end{itemize}
For pre-trained word embeddings, we apply the spectral algorithm of \newcite{stratos2015model} on a 2015 Korean Wikipedia dump
to induce 285,933 embeddings of dimension 100.

\paragraph{Parsing accuracy}
Table~\ref{tab:result} shows the main result.
The baseline test LAS of the original cross-lingual parser of McDonald13 is 55.85.
Yara achieves 85.17 with predicted POS tags and 88.61 with gold POS tags.
The basic BiLSTM model of K\&G16 obtains 82.77 with pre-trained word embeddings (78.95 without).
The stack LSTM parser is comparable to K\&G16 at the word level (Dyer15), but it performs significantly better at the character level (Ballesteros15) reaching 86.25 test LAS.

We observe decisive improvement when we incorporate sub-lexical information into the parser of K\&G16.
In fact, a \textit{strictly} sub-lexical parser using only jamos or characters clearly outperforms its lexical counterpart
despite the fact that the model is drastically smaller (e.g., 90.77 with $500 \by 100$ jamo embeddings vs 82.77 with $298115 \by 100$ word embeddings).
Notably, jamos alone achieve 91.46 which is not far behind the best result 92.31 obtained by using word, character, and jamo units in conjunction.
This demonstrates that our compositional architecture learns to build effective representations of Korean characters and words for parsing
from a minuscule set of jamo letters.

\section{Discussion of Future Work}
\label{sec:discussion}

We have presented a natural sub-character architecture to model the unique compositional orthography of the Korean language.
The architecture induces word-/sentence-level representations from a small set of phonetic units called jamo letters.
This is enabled by efficient and deterministic Unicode decomposition of characters.

We have focused on dependency parsing to demonstrate the utility of our approach as an economical and effective way to combat data sparsity.
However, we believe that the true benefit of this architecture will be more evident in speech processing as jamo letters are definitions of sound in the language.
Another potentially interesting application is informal text on the internet.
Ill-formed words such as \begin{CJK}{UTF8}{mj}ㅎㅎㅎ\end{CJK} (shorthand for \begin{CJK}{UTF8}{mj}하하하\end{CJK}, an onomatopoeic expression of laughter)
and \begin{CJK}{UTF8}{mj} ㄴㄴ\end{CJK} (shorthand for \begin{CJK}{UTF8}{mj}노노\end{CJK}, a transcription of \texttt{no no})
are omnipresent in social media. The jamo architecture can be useful in this scenario, for instance by correlating
\begin{CJK}{UTF8}{mj}ㅎㅎㅎ\end{CJK} and \begin{CJK}{UTF8}{mj}하하하\end{CJK} which might otherwise be treated as independent.

\section*{Acknowledgments}
The author would like to thank Lingpeng Kong for his help with using the DyNet library,
Mohammad Rasooli for his help with using the Yara parser,
and Karen Livescu for helpful comments.

\bibliography{korean}
\bibliographystyle{emnlp_natbib}

\clearpage\end{CJK}

\end{document}